\documentclass[letterpaper, 10 pt, conference]{ieeeconf}  

\IEEEoverridecommandlockouts                              

\usepackage{mathrsfs}
\usepackage{graphicx}
\usepackage{stfloats}
\usepackage{amsmath}
\usepackage{cases} 
\usepackage{overpic}
\usepackage{color}
\usepackage[T1]{fontenc}
\usepackage{aecompl}
\usepackage[skip=1ex]{caption}
\usepackage{booktabs,tabularx}
\usepackage{siunitx, cite,hyperref}
\newcolumntype{C}{>{\centering\arraybackslash}X}

\usepackage{threeparttable}

\title{\LARGE \bf
Learning Bifunctional Push-grasping Synergistic Strategy \\for Goal-agnostic and Goal-oriented Tasks
}

\author{Dafa Ren$^{*}$, Shuang Wu$^{**}$, Xiaofan Wang$^{*}$, Yan Peng$^{*}$ and Xiaoqiang Ren$^{*}$
	\thanks{$*$: School of Mechatronic Engineering and Automation, Shanghai University, Shanghai, 200444, China. {\tt\small Emails: (dafaren, xfwang, pengyan, xqren)@shu.edu.cn}}	
	\thanks{$**$: Huawei Noah's Ark Lab, {\tt\small Email: wushuang.noah@huawei.com} }%
	}%

\begin{document}

\maketitle
\thispagestyle{empty}
\pagestyle{empty}

\begin{abstract}
	Both goal-agnostic and goal-oriented tasks have practical value for robotic grasping: goal-agnostic tasks target all objects in the workspace, while goal-oriented tasks aim at grasping pre-assigned goal objects. However, most current grasping methods are only better at coping with one task. 
	In this work, we propose a bifunctional push-grasping synergistic strategy for goal-agnostic and goal-oriented grasping tasks. Our method integrates
	pushing along with grasping to pick up all objects or pre-assigned goal objects with high action efficiency depending on the task requirement.
	We introduce a bifunctional network, which takes in visual observations and outputs dense pixel-wise maps of $Q$ values for pushing and grasping primitive actions, to increase the available samples in the action space. 
	Then we propose a hierarchical reinforcement learning framework to coordinate the two tasks by considering the goal-agnostic task as a combination of multiple goal-oriented tasks. 
	To reduce the training difficulty of the hierarchical framework, we design a two-stage training method to train the two types of tasks separately.
	We perform pre-training of the model in simulation, and then transfer the learned model to the real world without any additional real-world fine-tuning.
	Experimental results show that the proposed approach outperforms existing methods in task completion rate and grasp success rate with less motion number. Supplementary material is available at \url{https://github.com/DafaRen/Learning\_Bifunctional\_Push-grasping\_Synergistic\_Strategy\_for\_Goal-agnostic\_and\_Goal-oriented\_Tasks}.
	
\end{abstract}

\section{Introduction}

\begin{figure}
	\centering
	\includegraphics[width=0.8\columnwidth]{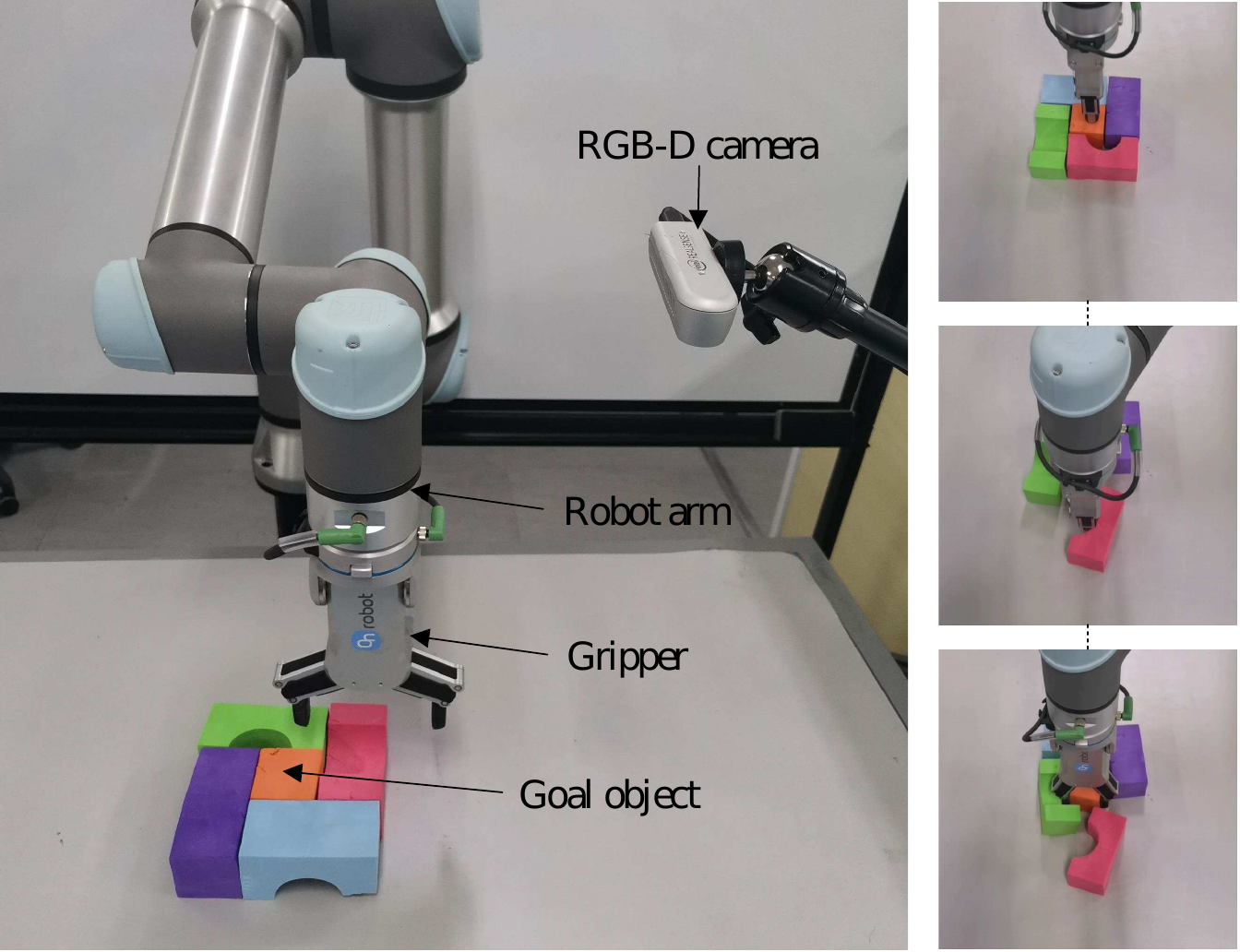}	
	\caption{\textbf{Example configuration of goal-oriented grasping tasks.} Obstacles around the goal object, in a cluttered environment, impede a successful grasp, which requires pre-grasping actions that can scatter the objects to facilitate grasping. Our system is able to pick up the goal object by integrating pre-grasping actions like pushing along with grasping.}
	\label{goal-environment}
\end{figure}

Robotic grasping task is one of the basic tasks of robotic manipulation, which is the key to robot-environment interactions. 
Grasping tasks can be divided into two groups depending on their target objects: goal-agnostic tasks and goal-oriented tasks.
Goal-agnostic tasks target all objects in the workspace, while goal-oriented tasks aim at grasping pre-assigned goal objects.

Despite the practical value of both goal-agnostic tasks and goal-oriented ones, most current grasping methods are only better at coping with one task.
One major challenge for grasping tasks is the existence of clutter in the environment, which impedes a successful grasp. One way to address this issue is to learn synergies between pushing and grasping actions~\cite{zeng2018learning, 8968042, 9165109}. However, most of these strategies are only applicable to relatively simple goal-agnostic tasks due to the limitations of network architectures and training method. While some strategies are able to grasp the designated target by incorporating goal object information~\cite{yang2020deep, xu2021efficient}, the action strategy is not efficient for goal-agnostic tasks. 

In this work, we propose a high action-efficient bifunctional push-grasping synergistic strategy through deep reinforcement learning for goal-agnostic and goal-oriented grasping tasks. We first train our policies end-to-end with a bifunctional network that takes in goal-agnostic visual observations of the scene and outputs dense pixel-wise maps of $Q$ values for pushing and grasping primitive actions. By leveraging the modular design and skip connections, we significantly increase the available samples in the action space.

We consider goal-oriented tasks as subtasks of a goal-agnostic task, which is different from existing methods that treat these two types of tasks as irrelevant. We model the goal-oriented task as a hierarchical reinforcement learning problem. The high-level controller selects the subtasks corresponding to the goal object in the goal-agnostic $Q$ map. The low-level controller instructs the robot to execute the actions with the maximum $Q$ value in the subtask region. 

To reduce the training complexity, we divide our system into two stages. The goal-agnostic task training stage focuses on precise grasping ability, while the goal-oriented task training stage is dedicated to learning synergy. To this end, we design different training environments and reward functions for these two stages.

In summary, our contributions are as follows:

\begin{itemize}
	\item We introduce a bifunctional network to output more accurate high-resolution $Q$ value predictions, which copes with both goal-agnostic and goal-oriented grasping tasks and improves the synergy.	
	\item We propose a hierarchical reinforcement learning framework to unify the two tasks by considering the goal-agnostic task as a combination of multiple goal-oriented tasks.	
	\item We design a two-stage training method to train the two tasks separately, which improves the sample efficiency and grasp success rate.
\end{itemize}

We perform several experiments and ablation studies in both simulation and the real world to demonstrate the effectiveness of our approach. The results show that the model transferred from simulation to the real world without fine-tuning can be applied to both goal-agnostic and goal-oriented tasks. Furthermore, our system performs higher task completion and grasp success rate with less motion number in both tasks.
\section{Related Work}

Robotic grasping techniques are active in robotics research. Classic analytical grasping solutions find stable force-closure for known objects by utilizing prior knowledge of object shapes, poses,  and dynamics that is difficult to know for novel objects in unstructured environments~\cite{bohg2013data, 7989165, liang2019pointnetgpd}. More recent data-driven methods explore model-agnostic grasping policies that directly link visual data to candidate grasps~\cite{zeng2018robotic, 9340777, mahler2017dex}.

Handling clutter is a major challenge today. To mitigate collisions from clutter, pre-grasping actions, such as pushing, have been introduced as primitive actions~\cite{deng2019deep, RN1, 9561073}. Zeng \textit{et al.}~\cite{zeng2018learning} proposed a $Q$-learning framework VPG to learn the complementary pushing and grasping policy, which copes with the clutter by executing complex sequential manipulations of objects in unstructured scenarios. Berscheid \textit{et al.}~\cite{8968042} introduced shifting actions to enable future grasps and bypassed sparse rewards by making shifting directly dependent on grasping. These methods target all objects in the workspace but cannot designate grasping objects. Therefore, these methods are only applicable to relatively simple goal-agnostic tasks, limiting the application scenarios of these methods.

In contrast to the goal-agnostic push-grasping described above, the goal-oriented push-grasping~\cite{yang2020deep, xu2021efficient, 8461041, 9197101} in cluttered scenes is more difficult, which needs to separate the goal object from its surroundings before the goal object is graspable. Kiatos and Malassiotis~\cite{8793972} proposed a push strategy to separate the target from its surrounding clutter through reinforcement learning. Sarantopoulos \textit{et al.}~\cite{9196647} employed a modular design to improve the convergence rate.
Yang \textit{et al.}~\cite{yang2020deep} trained a push-grasp strategy in a critic-policy
format to grasp initially invisible target objects. Xu \textit{et al.}~\cite{xu2021efficient} formulated the goal-oriented push-grasping as a hierarchical reinforcement learning problem. The grasp net serves as a discriminator to guide the training of the push net to improve synergy between pushing and grasping. However, when faced with a more cluttered environment, these methods suffer from reduced success rate and decreased action efficiency. 

Goal-agnostic and goal-oriented tasks have been extensively investigated. However, these two types of tasks are studied separately. Effective combination of both tasks is a relatively unexplored problem. Analogous to the above methods~\cite{zeng2018learning, yang2020deep, xu2021efficient}, our system utilizes a Fully Convolutional Network (FCN) as a function approximator to estimate the $Q$ function. However, there are several key differences that help our system outperform theirs. In their work, the network leverages spatial feature representations through direct bilinear upsampling, which considerably reduces the available samples in the action space. Besides, due to different design intentions, these methods cannot coordinate goal-agnostic and goal-oriented tasks. In contrast, we introduce a bifunctional network with accurate high-resolution $Q$ value predictions to increase the available samples. We leverage the hierarchical reinforcement learning framework to coordinate the two tasks. What's more, we design a two-stage training method to reduce the training complexity. Experiments demonstrate that our method is able to grasp the target more effectively with
less motion number in both tasks.

\section{Approach}

\begin{figure*}
	\centering
	\begin{overpic}[width=1\textwidth]{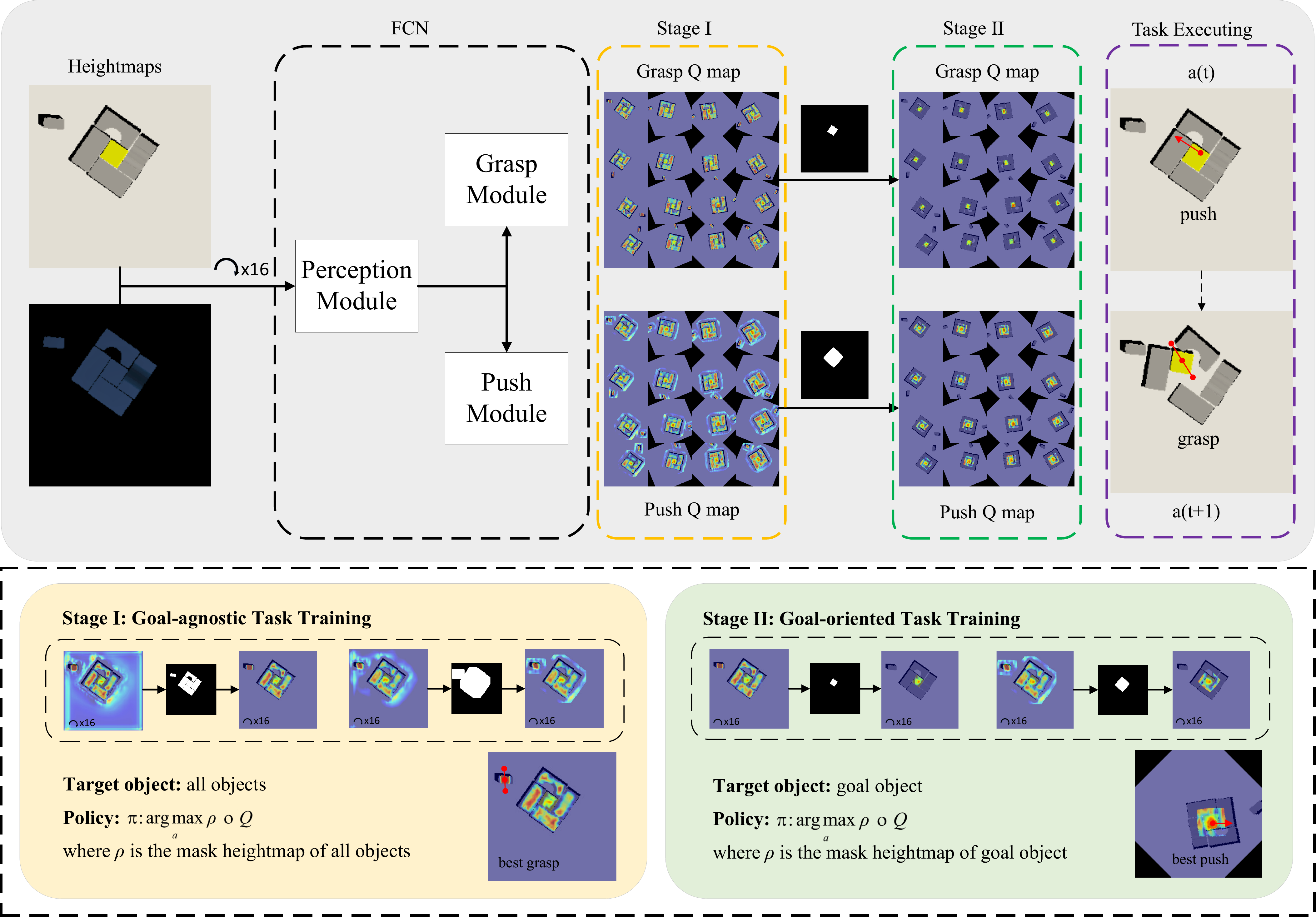}
		
	\end{overpic}
	\caption{\textbf{Overview.} The goal-agnostic visual observations captured by a statically mounted RGB-D camera are orthographically projected to construct heightmaps. Then, the heightmaps are rotated 16 times by an angle of $\pi/8$ radians and then fed into the FCN that outputs dense pixel-wise maps of $Q$ values. In the goal-agnostic task training stage, the robot will grasp the object that is most likely to succeed among all objects. In the goal-oriented task training stage, our robot arm performs a sequence of pushing and grasping actions to pick up the goal object in the mask region.}
	\label{overview}
\end{figure*}

We model the task of robotic grasping as a Markov decision process (MDP) with the state space $\mathcal{S}$, the action space $\mathcal{A}$, and the reward $\mathcal{R}$. 

We represent each state $s_t$ as RGB-D heightmaps, which are constructed by orthogonal projection of the visual 3D data observed by the statically mounted RGB-D camera. We represent the action space as a tuple $(x, y, z, \theta_i, \phi_j)$, where $(x, y, z)$ denotes the spatial coordinates of the gripper, $\theta$ is the rotation angle around the $z$-axis, $\phi$ corresponds to the top-down grasping action ($\phi_g$) or the straight pushing action ($\phi_p$).
As shown in Fig.~\ref{overview}, the RGB-D heightmaps are rotated 16 times by an angle of $\pi/8$ radians and then fed into the FCN. The rotation angle of the heightmap corresponds to the rotation angle of the gripper around the $z$-axis. The FCN outputs dense pixel-wise maps of $Q$ values with the same resolution as the heightmaps. Each pixel of the dense pixel-wise map represents a grasping or pushing action primitive at the corresponding 3D location, where each $Q$ value on that pixel represents the future expected reward of executing the manipulation primitive at the corresponding 3D location. 

In goal-agnostic tasks, we filter the $Q$ maps with the mask heightmap \cite{Liang} of all objects to obtain goal-agnostic $Q$ maps. The mask applied to the Push $Q$ map is the object mask obtained by the dilation operation. In goal-oriented tasks, we specify the goal object by the mask heightmap of the goal object. The goal-oriented $Q$ maps are generated from the Hadamard product of the goal mask and goal-agnostic $Q$ maps. The robot arm executes the action corresponding to the maximum $Q$ value according to the goal-oriented $Q$ maps.

To train the $Q$ function with this hierarchical reinforcement learning framework, we design a bifunctional FCN and propose a two-stage training pipeline.

\textbf{Stage \uppercase\expandafter{\romannumeral1}: Goal-agnostic Task Training.} In the first stage, we use a similar training approach as in [1]. The training goal of this stage is to make the network capable of performing goal-agnostic grasping tasks excellently. The target objects are all objects in the workspace. The system will grasp the object corresponding to the maximum $Q$ value in all objects. Push actions are only expected to be executed in extreme scenarios. In this stage, the system will learn precise grasping and a bit of push-grasping synergy.

\textbf{Stage \uppercase\expandafter{\romannumeral2}: Goal-oriented Task Training.} Unlike the previous stage, which is trained in relatively scattered scenarios, the network is trained in relatively cluttered scenarios. Our target is to learn to pick up the goal object in highly cluttered scenarios. This requires us to focus on training the synergy between grasping and pushing. Therefore, we design a goal-oriented pre-grasping action reward function and increase the frequency of pushing in highly cluttered scenarios. After the goal-oriented task training, the predicted grasp $Q$ values will be more robust. Our scheme improves the synergy between grasping and pushing and additionally enhances the ability of goal-agnostic grasping tasks.

\subsection{Network}

\begin{figure}
	\centering
	\includegraphics[width=1\columnwidth]{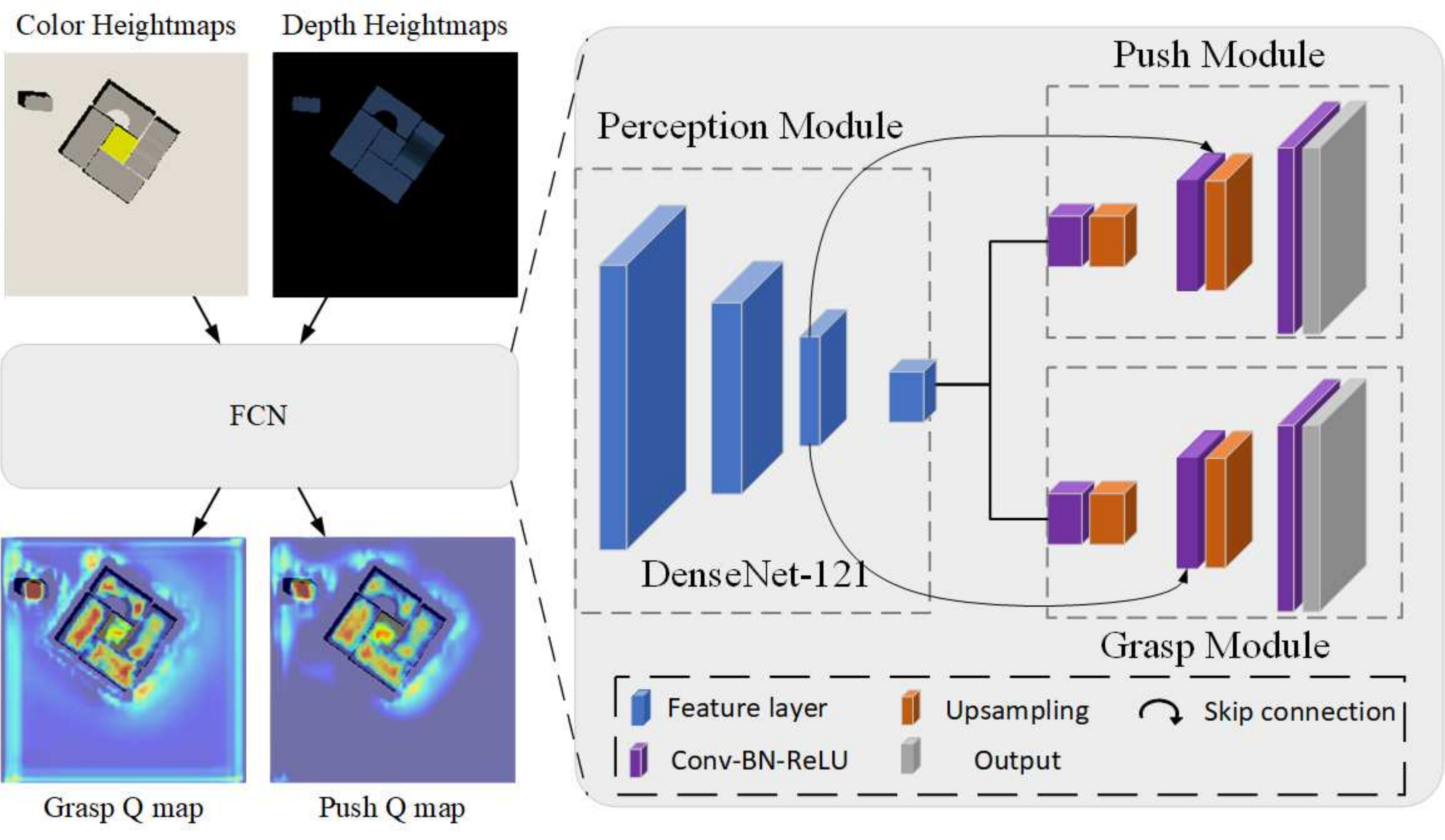}
	
	\caption{\textbf{Network architectures}} 
	\label{goal-network}
\end{figure}

We use Double DQN\cite{van2016deep} as the backbone of estimating the $Q$ functions by FCN~\cite{7298965}. The network takes 3D visual information as network input and outputs dense pixel-wise maps of $Q$ values with the same image size. Each $Q$ value at a pixel represents the future expected reward when the primitive is executed at the 3D location of the corresponding pixel. Our FCN consists of the following three parts:

1) \textbf{Perception module} consists of two DenseNet-121~\cite{8099726} networks that take RGB-D height maps as the input and output a spatial feature representation. This representation is shared as the input into the subsequent networks.

2) \textbf{Grasp module} establishes contact with the perception module through skip connections~\cite{peng2019pvnet} and outputs a dense pixel-wise map of $Q$ values as Grasp~$Q$~map for the grasping action.

3) \textbf{Push module} has the same structure as the grasp module and outputs a dense pixel-wise map of $Q$ values as Push~$Q$~map for the pre-grasping action.

In the network structure, we extend our previous work FLG~\cite{flg} by decoupling the grasping and pushing actions. FLG focuses on quickly learning to grasp and pre-grasp in goal-agnostic tasks. Its network outputs a dual-channel $Q$-value map. One channel corresponds to a grasping map and the other corresponds to a pushing map. This structure reduces the complexity of the network structure and speeds up the training process. However, it is challenging to tackle more complex scenarios in goal-oriented tasks. In this work, we enhance the prediction accuracy and stability of the network with a small increase in parameters through a modular design.

Both our and other similar methods \cite{yang2020deep,xu2021efficient} employ FCN to model the $Q$ function. However, there are two key differences that make our proposed method superior to theirs. 

First, the most significant difference between our network framework and theirs is whether the goal mask is taken as the input to the network. Their nets take as input the goal mask representations of the goal object. Noticeably, the work~\cite{xu2021efficient} designs a separate feature extraction network for the goal mask. Their methods ignore the connection between goal-oriented tasks and goal-agnostic ones by treating them as two completely unrelated tasks. In contrast to them, we argue that the goal-agnostic task can be considered as a combination of multiple goal-oriented tasks without taking the goal mask as input and designing a redundant feature extraction network. With an efficient network, the goal mask can serve as a selector that picks the subtask corresponding to the goal object through the goal-agnostic $Q$ maps. In addition, the method taking the target mask as input only outputs one $Q$ map for one object at a time, while our network outputs a $Q$ map for all objects.

Another noticeable difference is the processing of spatial feature representations. In their work, the spatial feature representation is directly bilinearly upsampled, considerably reducing the available samples in the action space. In theory, every pixel position could be sampled with the FCN. However, executing the action corresponding to the maximum $Q$ value results in selecting the position only at the resolution before upsampling. Unlike them, we concatenate spatial feature representations by a combination of skip connections and bilinear upsampling. The nets output more accurate high-resolution $Q$ value predictions, which leads to a higher grasp success rate and greater action efficiency.

\subsection{Goal-agnostic Task Training}
In this stage, the system focuses on precise grasping in the goal-agnostic task. We set up a relatively scattered scenario where 10 objects are randomly dropped into the workspace. We train 3000 steps for this stage. Due to the difference in reward functions and the relatively decentralized environment, the system is more biased to performing the grasping action. Pushing actions are executed only in rare scenarios. 

Both the reward function and the exploration policy are the same as in our previous work FLG\cite{flg}. We assign grasping action reward $R_g = 1$ if an object is successfully grasped. Pre-grasping action reward $R_p = 0.5$ is given for pushes that scatter the objects. The exploration policy introduces a mask function to prevent the robot from continuing exploration in object-less areas. The mask function filters the $Q$ map by masking the object-less areas.

\subsection{Goal-oriented Task Training}
The goal of this stage is to enable goal-oriented tasks through the synergy of grasping and pre-grasping actions.

\textbf{Rewards.} The reward function for goal-oriented tasks can be regarded as a refinement of the reward function for goal-agnostic tasks.

We define the reward $R_g$ for the goal-oriented grasping action as

\begin{equation}
	R_g =
	\begin{cases} 
		1,  & \mbox{if the goal is successfully grasped,} \\
		0, & \mbox{otherwise.}
	\end{cases}
\end{equation}

\begin{figure}
	\centering
	\includegraphics[width=1\columnwidth]{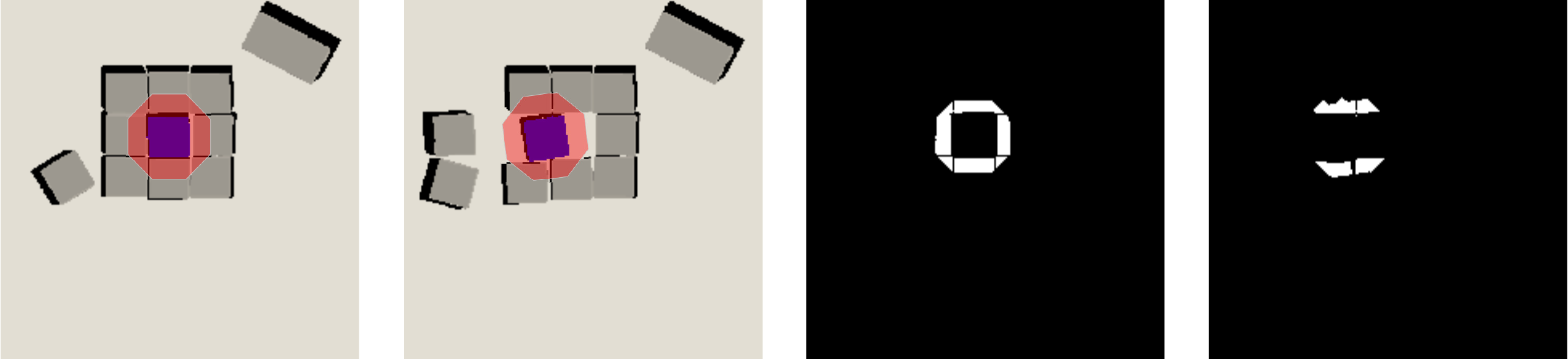}
	
	\caption{\textbf{Border occupancy visualization.} A potentially good push action will reduce the percentage of white pixels, which increases the free space around goal object \cite{yang2020deep}}.
	\label{goal-cqm}
\end{figure}

The purpose of the goal-oriented pre-grasping action is to make more space around the goal object by scattering the goal object and obstacles by pushing them away. To define the change in space, we first construct the  the mask of target border $m$ from the difference before and after the dilation of the target mask (the light red area in Fig.~\ref{goal-cqm}). Then the border occupancy value $m_v$ is defined as the number of pixels in $m$ above the workspace (the white area in Fig.~\ref{goal-cqm}). Given the difference in the size of the goal objects, we define the change in the free space around the target object $\eta$ as the change in border occupancy ratio $m_r$~\cite{yang2020deep}. More formally,
	
\begin{align*}
	m_r(t) &= \frac{m_v(t)}{m(t)}, \\
	\eta(t) &= m_r(t) - m_r(t+1).
\end{align*}

Push is effective for increasing the free space. Accordingly, we define the reward $R_p$ for goal-oriented pre-grasping action as

\begin{equation}
	R_p (s) =
	\begin{cases} 
		0.5,  & \mbox{if $\eta>\tau$,} \\
		0, & \mbox{else,}
	\end{cases}
\end{equation}
where $\tau$ is the preset threshold value.

\textbf{Training details.} To improve synergy, we set up a relatively cluttered scenario where seven candidate goal objects and 23 obstacles are randomly dropped into the workspace. The robot performs a sequence of grasping and pre-grasping actions to pick up a pre-assigned goal object. Once the goal is successfully grasped, a new goal is assigned until no more candidate goal objects are available in the workspace, at which point objects are again randomly dropped. In addition, we expect more pre-grasping action data to train the synergistic strategy in goal-oriented tasks. To increase the frequency of pre-grasping actions, the agent will perform a pre-grasping action if the maximum $Q$ value within the goal mask is below some threshold \cite{xu2021efficient}. We train 5000 steps for this stage.

Goal masks are generated by the semantic segmentation module~\cite{yang2020deep}. Our models are trained with an AMD 3970X processor and an NVIDIA RTX 2080Ti. Our system uses prioritized experience replay~\cite{schaul2015prioritized}, the exploration policy employs $\epsilon$-greedy strategy, and the future discount $\gamma$ is constant at 0.5. We train the FCN with the Huber loss function and Adam optimizer.

\section{Experiments}
In this section, we execute a series of experiments to evaluate our approach. The goals of the experiments are four-fold: 1) to compare our proposed method with other baseline alternatives, 2) to show the significance of the novel network architecture and the two-stage training approach, 3) to investigate whether our approach is effective for both goal-oriented tasks and goal-agnostic tasks, 4) to test whether our model can be successfully transferred from simulation to the real world without any fine-tuning.

\subsection{Baseline Methods}

We compare the performance of our system to the following baseline methods:

\textbf{Grasping the Invisible (GI)} is a target-oriented method utilizing DQN to train a push-grasp strategy with the goal of grasping initially invisible target objects~\cite{yang2020deep}. The method splits the problem into exploration and coordination subtasks depending on whether the target is detected by the segmentation module. If the target is invisible, the agent performs pushing to explore it. Once the target is found, the agent coordinates pushing and grasping to pick it up.

\textbf{Efficient Push-grasping (EPG)} is a goal-oriented push-grasping synergy strategy that formulates the synergy strategy as a hierarchical reinforcement learning problem~\cite{xu2021efficient}. For high-level control, a grasp net as a discriminator evaluates graspable probability with predicted grasp $Q$ values. For low-level control, a push net as a generator alters the graspability probability. If the maximum grasp $Q$ value exceeds a threshold, the robot will execute a grasp action, and activate a pushing action otherwise. 

\subsection{Evaluation Metrics}

We evaluate the methods with a series of test cases where the robot must face severely cluttered scenarios to pick up the goal object. For each test, we execute $n$ runs ($n=30$ in simulation and $n=15$ in the real world) and evaluate performance with three metrics same as those in \cite{xu2021efficient}:

\begin{itemize}
	\item \textbf{Completion}: the average percentage of completion rate over $n$ runs. The task is determined to be completed when the robot successfully grasps the goal object without 10 consecutive failed attempts.
	\item \textbf{Grasp success rate}: the average percentage of goal grasp success rate per completion.
	\item \textbf{Motion number}: the average number of motions per completion. Motion number is inversely proportional to	action efficiency.
\end{itemize}

\subsection{Simulation Experiments}

\begin{figure}
	\centering
	\includegraphics[width=0.6\columnwidth]{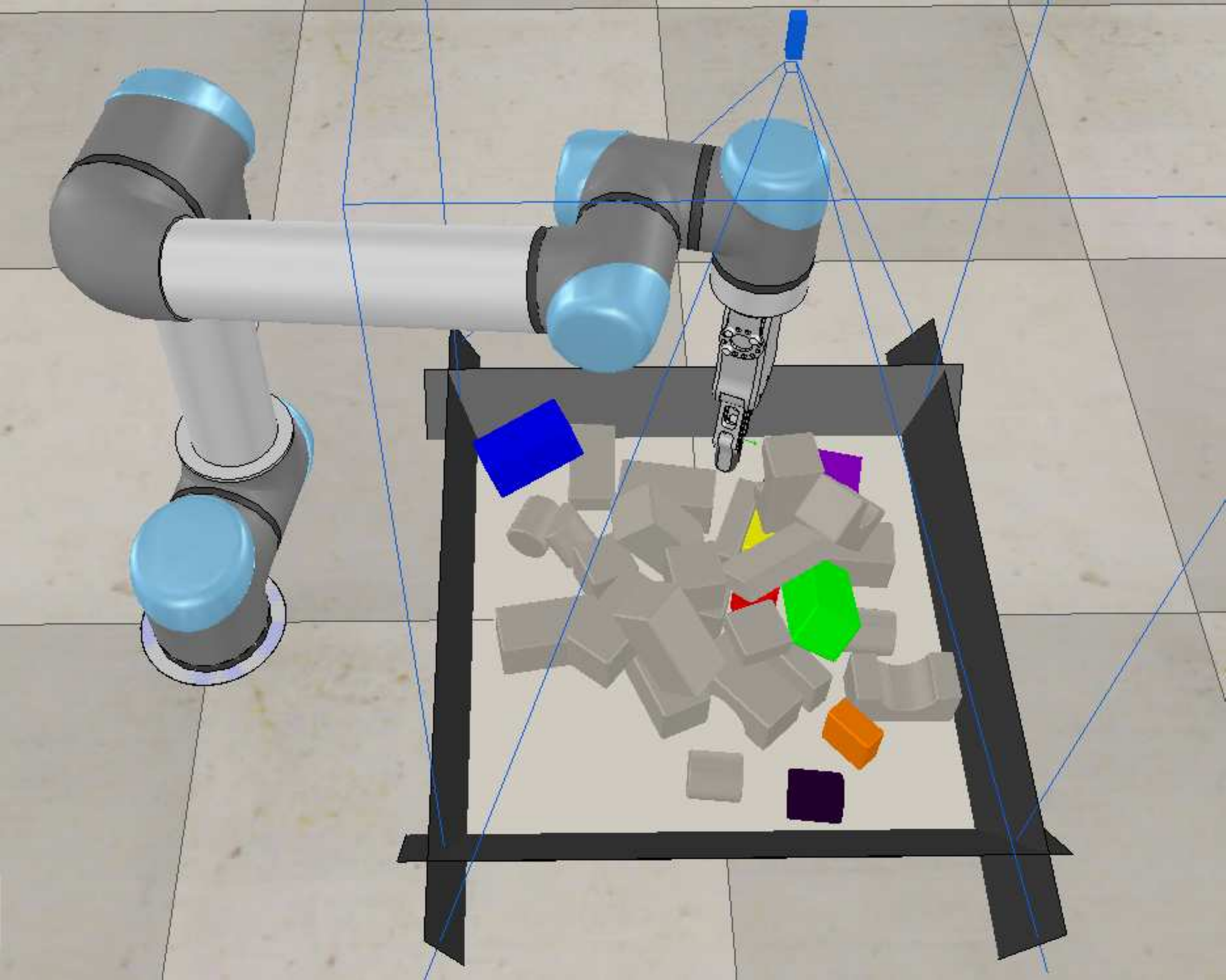}
	
	\caption{\textbf{Simulation environment.} Models are trained in scenarios with random arrangements of 7 candidate goal objects and 23 obstacles.}
	\label{sim-environment}
\end{figure}

Our simulation environment is similar to \cite{zeng2018learning}, which consists of a UR5 robot arm
and an RG2 gripper in CoppeliaSim (shown in Fig.~\ref{sim-environment}).

\subsubsection{Goal-oriented Tasks}
\textbf{Comparisons to Baselines.} We first compare the performance of our method with the baselines on random arrangements where 30 objects are randomly dropped into the workspace. Random arrangements are similar to the training scenarios, except that they contain one goal object and 29 obstacles. We observe that our method outperforms both baseline methods in task completion rate and goal grasp success rate with less motion number (see Table.~\ref{goal-test-r}). We speculate that the poor performance of the \textbf{GI} policy is due to the aggressive grasping policy and the outputs with lower actual resolution. When the edge of the goal object is not obstructed, this policy tends to grasp. However, a poor network limits it to execute a suboptimal action. While the \textbf{EPG} policy achieves higher completion and grasp success rates than the \textbf{GI} policy, the average motion number is still higher than ours. This suggests that the \textbf{EPG} policy improves synergy over the \textbf{GI} policy, but still lacks accuracy. This meets our expectation. Since the \textbf{EPG} policy actually samples less, it can only execute longer push-grasping sequences to pick up the goal object.

\begin{table}[htbp]  
	\centering 
	\caption{\label{goal-test-r}SIMULATION RESULTS ON RANDOM ARRANGEMENTS}   
	\setlength{\tabcolsep}{2mm}{
		\begin{tabular}{lcccc}    
			\toprule    
			&Method       & Completion       & Grasp Success       & Motion Number   \\
			\midrule
			&GI~\cite{yang2020deep}    & 96.7$\%$     & 54.6$\%$       & 5.37                         \\
			&EPG~\cite{xu2021efficient}      & 97.8$\%$    & 90.0$\%$     & 4.82                             \\
			&\textbf{Ours}       & \textbf{100$\%$}    & \textbf{94.4$\%$}          & \textbf{2.44}                  \\    
			\bottomrule   
	\end{tabular}}  
\end{table}

We also compare the performance difference between our method and the baselines on challenging arrangements which involve 10 test cases with adversarial structures. In each test case, the goal object is either placed closely side-by-side with the obstacle or surrounded by it, and even the optimal grasping policy must de-clutter before it can successfully pick up the goal object. Results are shown in Table~\ref{goal-test-c}. We observe that our method outperforms both baseline methods across three metrics. The \textbf{GI} policy performs poorly in both grasp success rate and motion number metrics. Although the \textbf{EPG} policy performs better than the \textbf{GI} policy, it is still weaker than our method. In dense cluttered scenarios such as case 7 and 9, the \textbf{EPG} policy suffers from decreased grasp success rate and increased motion number.

The results on random and challenging arrangements demonstrate the feasibility and efficiency of our method.

\begin{figure}[htbp] 
	\centering
	\includegraphics[width=1\columnwidth]{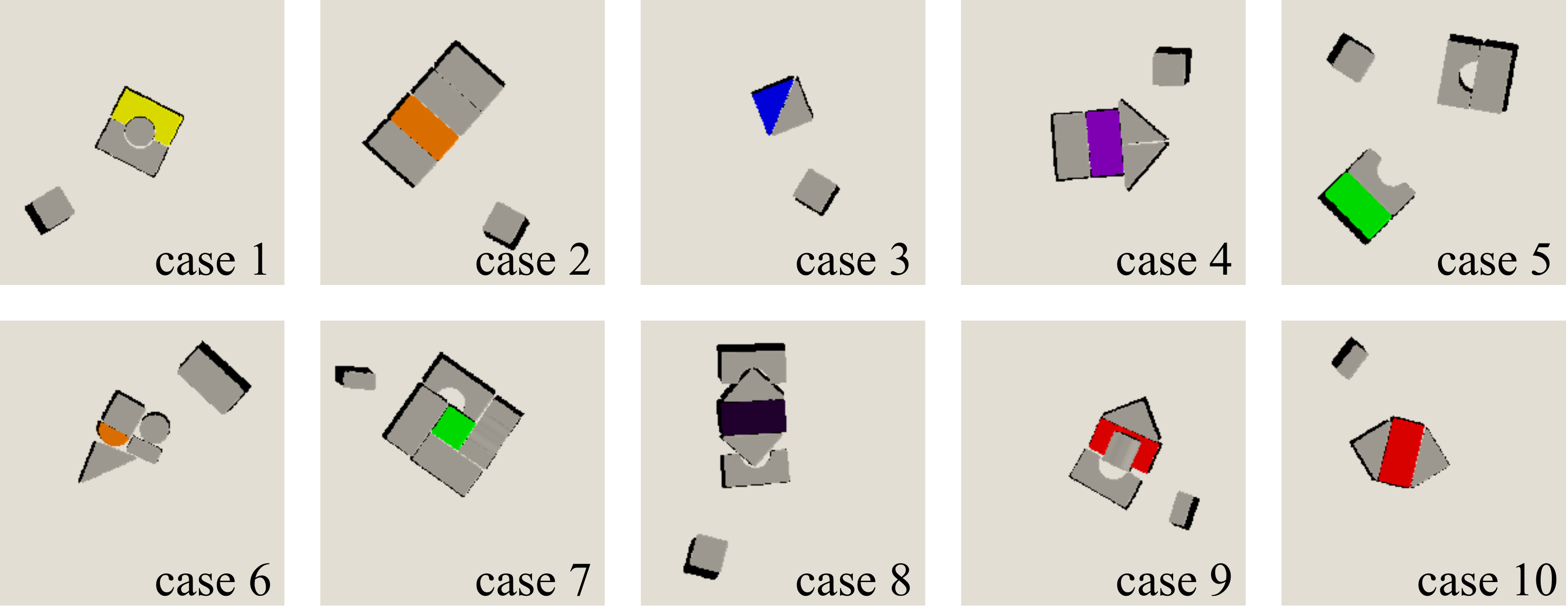}
	
	\caption{Challenging arrangements in simulation experiments which involve 10 designed scenes with adversarial clutter. The goal of each scene is the colored object.}
	\label{goal-test-case}
\end{figure}

\begin{table}[htbp]  
	\centering 
	\caption{\label{goal-test-c}SIMULATION RESULTS ON CHALLENGING ARRANGEMENTS}   
	\setlength{\tabcolsep}{2mm}{
		\begin{tabular}{lcccc}    
			\toprule    
			&Method       & Completion       & Grasp Success       & Motion Number   \\
			\midrule
			&GI~\cite{yang2020deep}    & 95.0$\%$     & 70.4$\%$       & 4.33                         \\
			&EPG~\cite{xu2021efficient}      & 99.0$\%$    & 90.0$\%$     & 2.77                    \\
			&\textbf{Ours}       & \textbf{100$\%$}    & \textbf{93.6$\%$}          & \textbf{2.54}                  \\    
			\bottomrule  

	\end{tabular}}  

\end{table}

\textbf{Ablation Studies.} We next run ablation studies to investigate: 1) whether the pre-grasping action can improve the grasp success rate, 2) whether the network architecture is simple and effective, 3) whether the two-stage training method can improve the sample efficiency. Table~\ref{ablation studies} reports the results of our method compared to the ablation methods on random arrangements.

\begin{table}[htbp]  
	\centering 
	\caption{\label{ablation studies}ABLATION STUDIES RESULTS}   
	\setlength{\tabcolsep}{1mm}{
		\begin{tabular}{lcccc}    
			\toprule    
			&Method       & Completion       & Grasp Success       & Motion Number   \\
			\midrule
			&Grasping-only    & 98.1$\%$     & 61.8$\%$       & 2.69                         \\
			&W/o two-stage training   & 100$\%$     & 84.7$\%$       & 3.40                         \\
			&\textbf{Ours}       & \textbf{100$\%$}    & \textbf{94.4$\%$}          &\textbf{2.44}                 \\    
			\bottomrule   
	\end{tabular}}  
\end{table}

We design the Grasping-only policy to verify the importance of the pre-grasping action. Grasping-only policy is a variant of our policy that uses the same state inputs as ours, but uses a single FCN to predict $Q$ values for grasping only, which means that its network contains only the perception module and the grasping module. As shown in Table~\ref{ablation studies}, our policy enables higher completion and grasp
success rates by associating planning with pushing and grasping. While the grasping-only policy uses fine-tuning to enable harder grasps, our policy can perform pre-grasping actions that facilitate grasping.

\begin{figure}
	\centering
	\includegraphics[width=0.9\columnwidth]{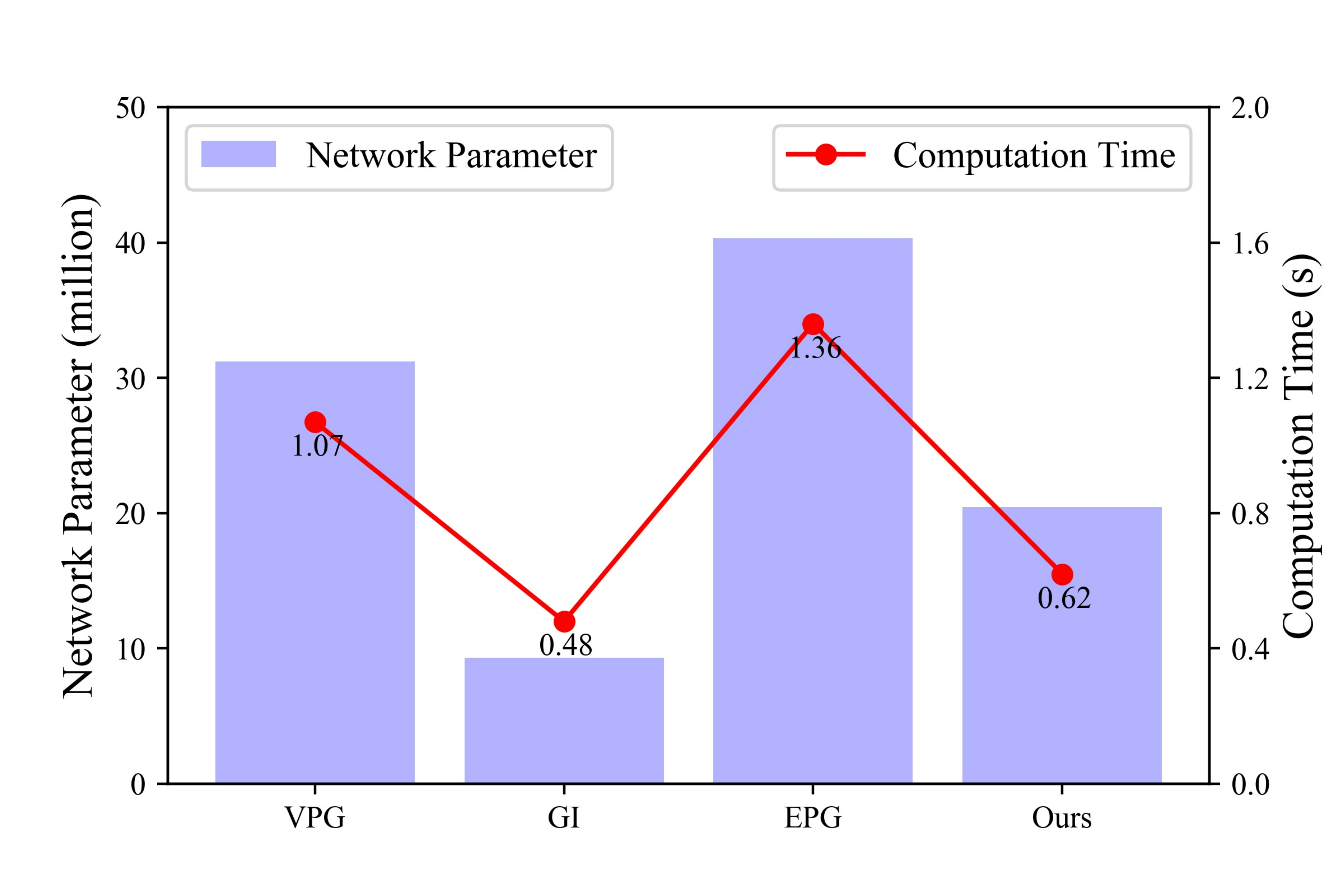}
	
	\caption{Comparison the number of network parameters and the $Q$ values computation time.} 
	\label{goal-net-compare}
\end{figure}

We then compare the number of network parameters and the $Q$ values computation time of our method with the other methods. Results are shown in Fig.~\ref{goal-net-compare}. Our network parameters and computation time are significantly less than \textbf{EPG} and \textbf{VPG}, but slightly more than \textbf{GI}. Our computation time (0.62 s) is half of \textbf{EPG}, but our performance surpasses them, which shows the simplicity and effectiveness of our network. Moreover, although \textbf{GI} has the least computation time, it does not mean that it has a high efficiency since its grasp success rate is much lower than ours.

\begin{figure}
	\centering
	\includegraphics[width=1\columnwidth]{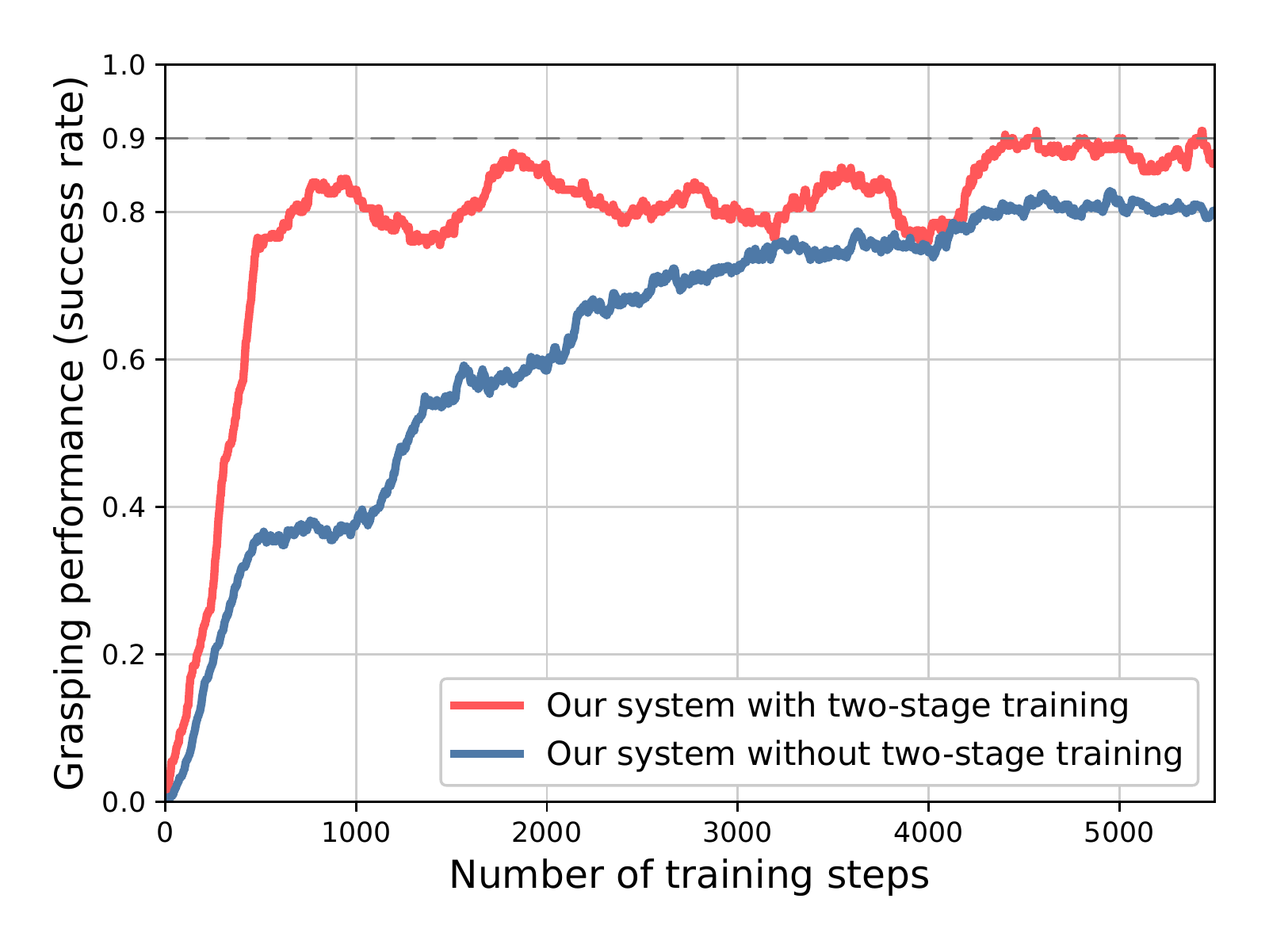}
	
	\caption{Comparing performance of our policies trained with and without two-stage training. The success rate of each step is the average percentage of goal grasp success rate of the first 200 steps of the current step.} 
	\label{goal-two-stage}
\end{figure}

We also train a version of the policy with only goal-oriented task training (``our system without two-stage training") and report its goal-oriented grasping performance versus training steps in Fig.~\ref{goal-two-stage}. Note that the training scene contains 30 objects instead of 10. From these curves, we see that our system with two-stage training improves its grasping performance at a faster pace early in training, which converges to 80$\%$ grasp success rate in just 650 action attempts. From Table~\ref{ablation studies}, we see that the system with two-stage training improves the grasp success rate by 10$\%$ and reduces the motion number by 1. This suggests that the two-stage training method can significantly improve the sample efficiency and performance. We speculate that this is because the two-stage training approach decomposes the task into two easier tasks. The goal-agnostic task training stage focuses on precise grasping ability, while the goal-oriented task training stage is dedicated to learning synergy.

\subsubsection{Goal-agnostic Tasks} We further investigate whether our approach is effective for goal-agnostic tasks. We first evaluate our method in goal-agnostic challenging arrangements with the same object shapes and locations as in Fig.~\ref{goal-test-case}, except that the colors of obstacles are replaced. In this experiment, we compare our approach with \textbf{VPG} and \textbf{EPG} and report the results in Table~\ref{goal-agnostic-c}. \textbf{EPG} beats \textbf{VPG} in task completion rate and grasping success rate, but the action efficiency of \textbf{EPG} is lower than that of \textbf{VPG}. This is because \textbf{EPG} is designed for goal-oriented tasks. Each each time an object is specified, \textbf{EPG} performs a series of push-grasp actions until that goal object is successfully grasped. This leads to many ineffective actions even if there are better-grasped objects in the process. Our method outperforms the other two methods in all metrics, demonstrating the generality of our method for both goal-oriented and goal-agnostic tasks.

\begin{table}[htbp]  
	\centering 
	\caption{\label{goal-agnostic-c}PERFORMANCE IN GOAL-AGNOSTIC CONDITION}   
	\setlength{\tabcolsep}{1mm}{
		\begin{tabular}{lcccc}    
			\toprule    
			&Method       & Completion       & Grasp Success       & Action Efficiency  \\
			\midrule
			&VPG~\cite{zeng2018learning}   & 82.7$\%$     & 77.2$\%$       & 60.1$\%$                         \\
			&EPG~\cite{xu2021efficient}      & 95.1$\%$    & 82.5$\%$     & 56.7$\%$                    \\
			&\textbf{Ours}       & \textbf{100$\%$}    & \textbf{90.3$\%$}          & \textbf{61.1$\%$}                  \\    
			\bottomrule   
	\end{tabular}}  
\end{table}

\subsection{Real-World Experiments}
In this section, we evaluate our system on a real
robot. Our real-world setup uses a UR5e robot with an RG2 gripper as the end effector. RGB-D images are captured from an Intel RealSense D415 mounted statically above the workspace. 

\subsubsection{Goal-oriented Tasks} We first test our system in goal-oriented tasks. Our test cases are kept the same with \cite{xu2021efficient} for fair comparisons, which consists of four random and challenging arrangements. We compare the performance of our method with the baselines. Results are shown in Table~\ref{real-goal-test-rc}. From these results, we see that our method outperforms the baseline in task completion rate and goal grasp success rate with less motion number. Note that the models of all three methods are transferred from simulation to the real world without any fine-tuning. This suggests that our policy can be effectively applied to the real world.

\begin{figure}
	\centering
	\includegraphics[width=1\columnwidth]{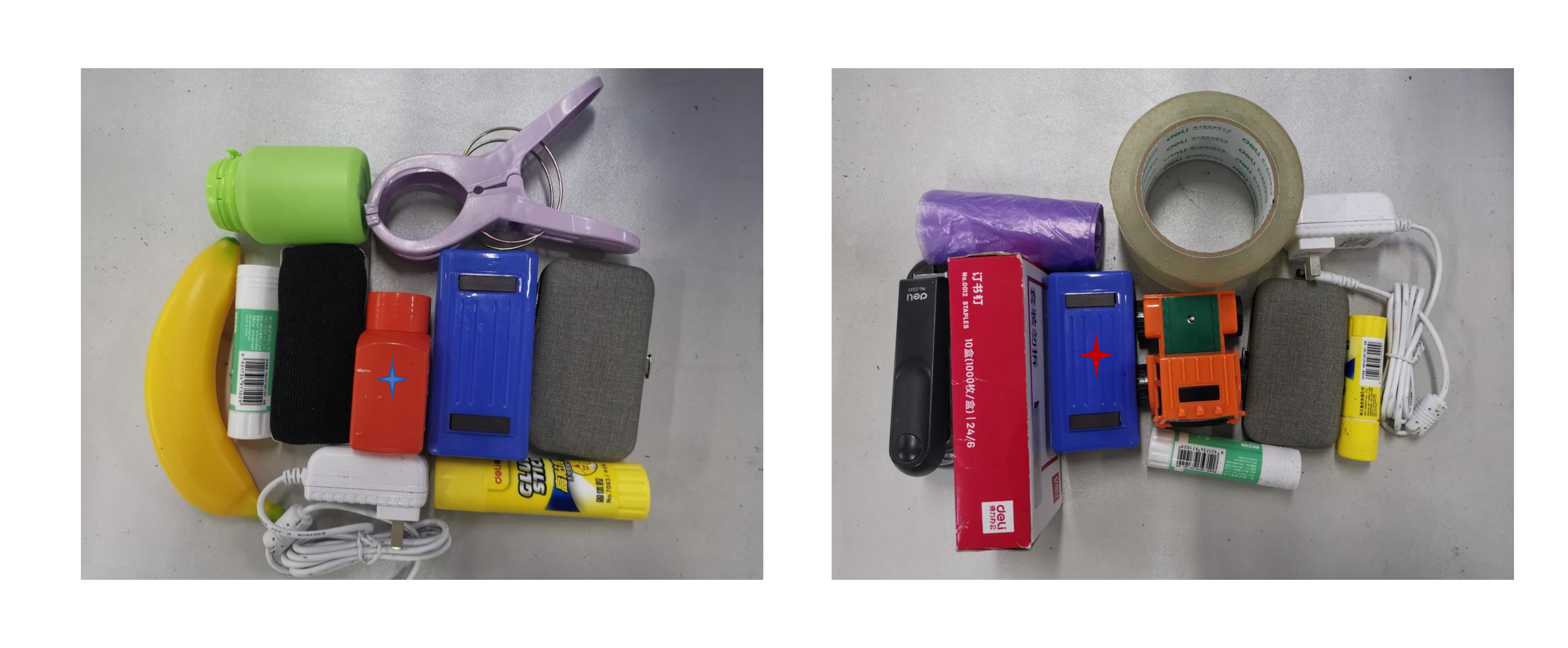}
	
	\caption{Example of real-world testing cases with novel objects for testing our systems ability to generalize to novel objects.} 
	\label{novel}
\end{figure}

We also test the ability to generalize to novel objects in a collection of real-world scenarios (shown in Fig.~\ref{novel}). The system achieved an average success rate of 89.4$\%$ with an average motion number of 2.43. Although our training set only has blocks, it is capable of generalizing to the normal objects with similar shapes.

\begin{table}[htbp]  
	\centering 
	\caption{\label{real-goal-test-rc}REAL-WORLD RESULTS IN GOAL-ORIENTED TASKS}   
	\setlength{\tabcolsep}{1.5mm}{
		\begin{threeparttable}
			\begin{tabular}{lccccccc}    
				\toprule    
				&Method       & \multicolumn{2}{c}{Completion}       & \multicolumn{2}{c}{Grasp Success}       & \multicolumn{2}{c}{Motion Number}   \\
				\midrule
				&Arrangement   &r &c &r &c &r &c \\
				\midrule
				&GI~\cite{yang2020deep}    & 86.7$\%$ & 85.0$\%$    & 75.2$\%$   & 70.3$\%$    & 6.92    & 6.81                         \\
				&EPG~\cite{xu2021efficient}   & 93.3$\%$    & 95.0$\%$  & 81.7$\%$   & 86.6$\%$    & 5.67     & 4.62                    \\
				&\textbf{Ours}    &\textbf{100$\%$}    & \textbf{100$\%$}  & \textbf{93.7$\%$}    & \textbf{97.1$\%$}      & \textbf{1.25}        & \textbf{2.23}                  \\    
				\bottomrule   
			\end{tabular}
			\begin{tablenotes} 
				\item[1] The second row of table represents arrangement type, where r and c correspond to random and challenging arrangements, respectively. 
			\end{tablenotes} 
	\end{threeparttable}}  
\end{table}

\subsubsection{Goal-agnostic Tasks} Finally, we test our system in goal-agnostic tasks on random arrangements, where 20 toys with varied shapes and colors in the workspace must be grasped. Our method outperforms other methods of reinforcement learning for goal-agnostic grasping tasks (see Table~\ref{real-goal-agnostic}). In addition, our model is tested directly without real-world training. This experiment suggests that our policy is robust on the real robot in goal-agnostic tasks.

\begin{table}[htbp]  
	\centering 
	\caption{\label{real-goal-agnostic}REAL-WORLD RESULTS IN GOAL-AGNOSTIC TASKS} 
	\setlength{\tabcolsep}{0.9mm}{
		\begin{threeparttable}
			\begin{tabular}{lccccccc}    
				\toprule    
    			Method    & Grasp Success    & Iteration Steps   & Test Items\\
    			\midrule
    			QT-opt~\cite{kalashnikov2018qt}    & 88$\%$             & 580k      &28                   \\
    			VPG~\cite{zeng2018learning}      & 68$\%$             & 2.5k    &20                          \\
    			Berscheid \textit{et al.}~\cite{8968042}    & 92$\%$            & 27.5k    &20                            \\
    			SPOT~\cite{9165109}      & 75$\%$           & 1k &20\\
    			FLG~\cite{flg}        & 94$\%$             & 2.5k       &20   \\
    			\textbf{Ours}                           & \textbf{95.5$\%$}             & \textbf{0}         &20                       \\    
    			\bottomrule   
			\end{tabular}
			\begin{tablenotes} 
				\item[1] Iteration Steps represents the number of training steps in the real world. Test Items corresponds to the number of objects in the scenario.
			\end{tablenotes} 
	\end{threeparttable}}  
\end{table}

\section{Conclusion}

In this work, we presented a hierarchical reinforcement learning framework to coordinate goal-agnostic grasping tasks and goal-oriented ones. We evaluated the performance of our system in both simulation and the real world. The experimental results indicated that our policy can learn synergy between pushing and grasping to accurately pick up all objects or pre-assigned goal objects in the clutter with high action efficiency.  Moreover, our system is feasible for practical deployment as the pre-trained model in the simulation achieved a considerably high success rate in the real world without fine-tuning.




%
%

%
%
%
%

\bibliographystyle{IEEEtran}
\bibliography{conf}
\end{document}